Ibrahim Alshubaily
Xiangliang Zhang
Directed Research
1 Aug 2019

# Speeding Up Neural Architecture Search By Using Performance Prediction

## Introduction

Neural networks are powerful models that have a remarkable ability to extract patterns that are too complex to be noticed by humans or other machine learning models. Neural networks are the first class of models that can train end-to-end systems with large learning capacity. However, we still have the difficult challenge of designing the neural network, which requires human experience and a long process of trial and error. As a solution, we can use a neural architecture search to find the best network architecture for the task at hand. Existing NAS algorithms generally evaluate the fitness of a new architecture by fully training from scratch, resulting in the prohibitive computational cost, even if operated on high-performance computers. In this paper, an end-to-end offline performance predictor is proposed to accelerate the fitness evaluation.

## Related Work

Barret Zoph and Quoc V. Le from the Google Brain team [15] introduced the idea of using a recurrent network to generate the model descriptions of neural networks and train this RNN with reinforcement learning to maximize the expected accuracy of the generated architectures on a validation set. Their paper showed promising results, inspired the work of this project, and provided great guidance.

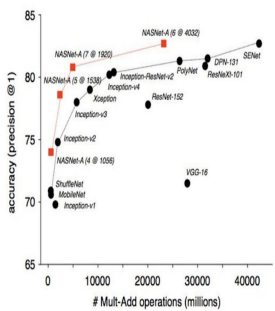
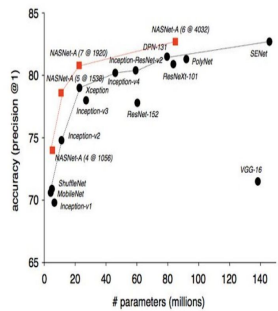

In "Large-Scale Evolution of Image Classifiers," presented at ICML 2017, an evolutionary process with simple building blocks and trivial initial conditions was introduced. The idea was to "sit back" and let evolution at scale do the work of constructing the architecture. Starting from very simple networks, the process found classifiers comparable to hand-designed models at the time.

To build on the great work done in this area, we employ performance prediction, in order to avoid fully training architectures from scratch. This addition has great potential because training the sampled architectures is computationally expensive and limits the number of architectures that can be explored.

# Methods

In this section, I will describe the work done in three essential components for neural architecture search. Which are the search space, search strategy, and performance estimation strategy.

*Search Space*

The search space defines which architectures can be represented in principle. Incorporating prior knowledge about properties well suited for a task can reduce the size of the search space and simplify the search. However, this also introduces a human bias, which may prevent finding novel architectural building blocks that go beyond the current human knowledge.

Google brain team [15], proposed a robust search space that consists of all possible directed acyclic graphs on V nodes, where each possible node has one of L labels, representing the corresponding operation. Two of the vertices are specially labeled as operation IN and OUT, representing the input and output tensors to the cell, respectively.

In order to limit the size of the space they propose the following restrictions:
Possible operations (L) = (3 × 3 convolution , 1 × 1 convolution, 3×3 max-pool). They also limit the number of vertices to be at most 7, and the number of edges to be at most 9.

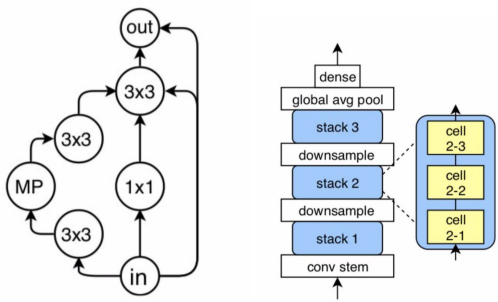

This cell design (left figure) had the highest validation accuracy on CIFAR-10 (94.5%). To achieve this result, the cell is stacked three times followed by a downsampling layer. This process is repeated three times, then a global average pooling and a dense layer are added at the end as shown in the right figure.

*Search Strategy*

The search strategy details how to explore the search space. It encompasses the classical exploration-exploitation trade-off, on the one hand, it is desirable to find well-performing architectures quickly, while on the other hand, premature convergence to a region of suboptimal architectures should be avoided. The two most popular algorithms for NAS are reinforcement learning - policy gradient, and regularized evolution.

<u>Reinforcement learning - policy gradient [15]:</u>

The list of tokens that the controller predicts can be viewed as a list of actions to design an architecture for a child network. At convergence, this child network will achieve an accuracy R on a held-out dataset. We can use this accuracy R as the reward signal and use reinforcement learning to train the controller.

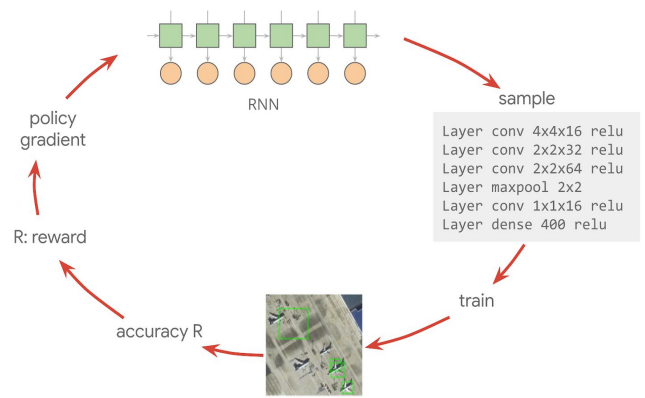

More concretely, to find the optimal architecture, we ask our controller to maximize its expected reward, represented by J(θc):

$$J(\theta_c) = E_{P(a_{1:T};\theta_c)}[R]$$

Since the reward signal R is non-differentiable, we need to use a policy gradient method to iteratively update θc.

REINFORCE rule from Williams (1992):

$$\nabla_{\theta_c} J(\theta_c) = \sum_{t=1}^{T} E_{P(a_{1:T};\theta_c)} \left[ \nabla_{\theta_c} \log P(a_t|a_{(t-1):1};\theta_c) R \right]$$

An empirical approximation of the above quantity is:

$$\frac{1}{m} \sum_{k=1}^{m} \sum_{t=1}^{T} \nabla_{\theta_c} \log P(a_t|a_{(t-1):1};\theta_c) R_k$$

Where m is the number of different architectures that the controller samples in one batch and T is the number of hyperparameters our controller has to predict to design a neural network architecture. The validation accuracy that the k-th neural network architecture achieves after being trained on a training dataset is Rk.

The above update is an unbiased estimate for our gradient but has a very high variance. In order to reduce the variance of this estimate we employ a baseline function:

$$\frac{1}{m} \sum_{k=1}^{m} \sum_{t=1}^{T} \nabla_{\theta_c} \log P(a_t|a_{(t-1):1};\theta_c) (R_k - b)$$

As long as the baseline function b does not depend on the current action, then this is still an unbiased gradient estimate. In this work, the baseline b is an exponential moving average of the previous architecture accuracies.

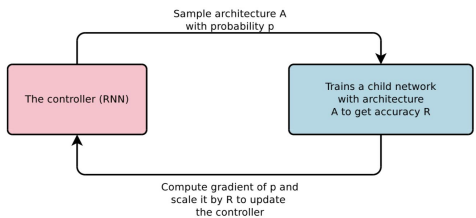

Regularized evolution [11]

**Algorithm 1** Aging Evolution

```
population ← empty queue          ▷ The population.
history ← ∅                       ▷ Will contain all models.
while |population| < P do         ▷ Initialize population.
    model.arch ← RANDOMARCHITECTURE()
    model.accuracy ← TRAINANDEVAL(model.arch)
    add model to right of population
    add model to history
end while
while |history| < C do            ▷ Evolve for C cycles.
    sample ← ∅                    ▷ Parent candidates.
    while |sample| < S do
        candidate ← random element from population
            ▷ The element stays in the population.
        add candidate to sample
    end while
    parent ← highest-accuracy model in sample
    child.arch ← MUTATE(parent.arch)
    child.accuracy ← TRAINANDEVAL(child.arch)
    add child to right of population
    add child to history
    remove dead from left of population    ▷ Oldest.
    discard dead
end while
return highest-accuracy model in history
```

The evolutionary method used keeps a population of P trained models throughout the experiment. The population is initialized with models with random architectures. After this, evolution improves the initial population in cycles. At each cycle, it samples S random models from the population, each drawn uniformly at random with replacement.

The model with the highest validation fitness within this sample is selected as the parent. A new architecture, called the child, is constructed from the parent by the application of a transformation called a mutation.

A mutation causes a simple and random modification of the architecture. Once the child architecture is constructed, it is then trained, evaluated, and added to the population. An interesting paper [3] talks about transferring the weights from the parent to the child to speed up the training. This idea was not pursued since the main focus of this paper is performance prediction.

### Performance Estimation Strategy:

The objective of NAS is typically to find architectures that achieve high predictive performance on unseen data. Performance Estimation refers to the process of estimating this performance: the simplest option is to perform a standard training and validation of the architecture on data, but this is unfortunately computationally expensive and limits the number of architectures that can be explored. Much recent research, therefore, focuses on developing methods that reduce the cost of these performance estimations.

Some performance estimation methods try to predict performance based on the validation accuracy in the first T epochs, other methods directly map architectures to performance, which is much more efficient if it works. Given enough training data, a recurrent neural network is a good model for mapping architectures to the final validation accuracy, due to its good performance and capability to handle variable sequence lengths (different CNN architectures vary in shape).

Predicting the final validation accuracy of architectures is a difficult task, it requires a large number of training samples. We can relax the problem, to reduce the number of samples required to train the RNN. Instead of predicting the final validation accuracy of architectures, we simplified the task to binary classification. Given an architecture, what is the likelihood that it will perform well. This simplification reduced the number of architectures required to train for the RNN. Besides reducing the number of required samples, we don't have to fully train samples, since our target label is one, if the architecture can achieve > %90 accuracy, and zero otherwise.

Learning to Teach [4] is very useful for speeding up the training of the performance predictor because L2T leads to a faster convergence which will reduce the time required for generating the training data. Additionally sharing parameters among similar architectures [7] can also reduce the time required for generating the training data.

## Results

The results of the experiment show that by using a reliable offline trained performance estimator, we can eventually obtain the same result with less than half the computational cost.

When applying REINFORCE the search converge after sampling ~900 architectures, the RNN requires 400 training samples to produce reliable predictions. However, this is not a hard requirement, the number of training samples for the performance estimator can be reduced further, one idea for reducing this number is to train an ensemble of performance predictors.

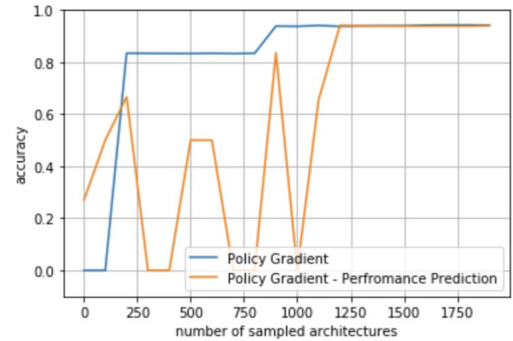

The performance predictor can be used to speed up reinforcement learning, instead of fully training an architecture from scratch to obtain the accuracy, which will be then used as a reward, we can simply use the performance predictor instead.

Similarly, The same performance predictor can be for the evolutionary search:

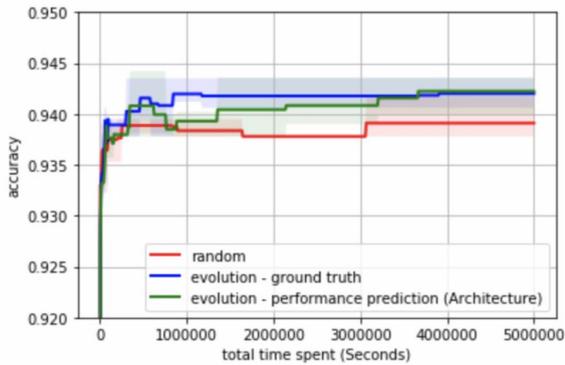

The x-axis represents the time needed according to the NASBENCH dataset to fully train all sampled architectures from scratch to obtain the validation accuracy. Note that this applies only to evolution - ground truth because in this algorithm the fitness function fully trains architectures from scratch to obtain the validation accuracy in order to evolve the population (very expensive). In evolution - performance prediction, we use the proposed RNN to instantly evaluate architectures. This figure shows that a reliable performance prediction model can provide a very significant speedup without losing quality because this method will also find the global optimal architecture eventually.

Evolution using true performance converges after 1 million seconds. According to the team that generated NASBENCH [15], that's equivalent of roughly 1200 evaluated architectures. One the other hand, evolution using the performance estimator require training 400 samples only. In this experiment, the performance estimator provided a 3X speedup.

## Conclusion

This paper presented the effectiveness of performance prediction to improve the efficiency of neural architecture search. The performance predictor can provide a significant speedup without loss in quality, regardless of which search algorithm is used.

To build upon the work done here, we can combine the strengths of reinforcement learning and evolutionary learning together. This method is known as Reinforced Evolutionary Neural Architecture Search, which is an evolutionary method with a reinforced mutation. This method integrates reinforced mutation into an evolution algorithm for neural architecture search, in which a mutation controller is introduced to learn the effects of slight modifications and make mutation actions. The reinforced mutation controller guides the model population to evolve efficiently [3].

In terms of a future direction, AutoML has much more to offer besides neural architecture search. Such as learning data augmentation policies, model scaling, and hyperparameter optimization. Additionally, NAS has been mainly focused on computer vision, it is important to go beyond image classification problems by applying NAS to less explored domains. Such as image restoration, semantic segmentation , transfer learning, machine translation, and multi-objective problems.

## References

1. Adam, George, and Jonathan Lorraine. Understanding Neural Architecture Search Techniques. 31 Mar. 2019, arxiv.org/abs/1904.00438.

2. Burges, Chris, and Tal Shaked. "Learning to Rank Using Gradient Descent." ACM Digital Library, ACM, dl.acm.org/citation.cfm?id=1102363.

3. Chen, Yukang, et al. Reinforced Evolutionary Neural Architecture Search. 10 Apr. 2019, arxiv.org/abs/1808.00193.

4. Fan, Yang, et al. "Learning to Teach." ArXiv.org, 9 May 2018, arxiv.org/abs/1805.03643.

5. Jiang, Yang, et al. "Neural Architecture Refinement: A Practical Way for Avoiding Overfitting in NAS." ArXiv.org, 7 May 2019, arxiv.org/abs/1905.02341.

6. Klein, Aaron, and Stefan Falkner. "Learning Curve Prediction with Bayesian Neural

Networks." Venues, ICLR 2017, 4 Nov. 2016, openreview.net/forum?id=S11KBYclx.

7. Pham, Hieu, et al. "Efficient Neural Architecture Search via Parameter Sharing." ArXiv.org, 12 Feb. 2018, arxiv.org/abs/1802.03268.

8. Saltori, Cristiano, et al. "Regularized Evolutionary Algorithm for Dynamic Neural Topology Search." ArXiv.org, 15 May 2019, arxiv.org/abs/1905.06252.

9. Singh, Prabhant, et al. "A Study of the Learning Progress in Neural Architecture Search Techniques." ArXiv.org, 18 June 2019, arxiv.org/abs/1906.07590.

10. Sun, Yanan, et al. "Evolving Deep Convolutional Neural Networks for Image Classification." ArXiv.org, 10 Mar. 2019, arxiv.org/abs/1710.10741.

11. Sun, Yanan, et al. "Surrogate-Assisted Evolutionary Deep Learning Using an End-to-End Random Forest-Based Performance Predictor." IEEE Journals & Magazine, 24 June 2019, ieeexplore.ieee.org/document/8744404.

12. Tan, Mingxing, and Quoc V. Le. "EfficientNet: Rethinking Model Scaling for ConvolutionalNeural Networks." ArXiv.org, 10 June 2019, arxiv.org/abs/1905.11946.

13. Wang, Tianyang, et al. "Data Dropout: Optimizing Training Data for Convolutional Neural Networks." ArXiv.org, 7 Sept. 2018, arxiv.org/abs/1809.00193.

14. Yang, Chuanguang, et al. "EENA: Efficient Evolution of Neural Architecture." ArXiv.org, 20 May 2019, arxiv.org/abs/1905.07320.

15. Ying, Chris, et al. "NAS-Bench-101: Towards Reproducible Neural Architecture Search." ArXiv.org, 14 May 2019, arxiv.org/abs/1902.09635.

16. Zoph, Barret, and Quoc V. Le. "Neural Architecture Search with Reinforcement Learning." ArXiv.org, 15 Feb. 2017, arxiv.org/abs/1611.01578.

17. Zoph, Barret, et al. "AutoAugment: Learning Augmentation Policies from Data." ArXiv.org, 11 Apr. 2019, arxiv.org/abs/1805.09501.